\documentclass{article} %
\usepackage{iclr2026_conference,times}

\usepackage{amsmath,amsfonts,bm}

\def\eqref#1{equation~\ref{#1}}

\def\1{\bm{1}}

\DeclareMathAlphabet{\mathsfit}{\encodingdefault}{\sfdefault}{m}{sl}
\SetMathAlphabet{\mathsfit}{bold}{\encodingdefault}{\sfdefault}{bx}{n}

\usepackage{hyperref}
\usepackage{url}
\usepackage{graphicx}
\usepackage{booktabs}
\usepackage{subcaption}
\usepackage{wrapfig}
\usepackage{listings}
\usepackage{xcolor}
\usepackage[export]{adjustbox}
\usepackage[T1]{fontenc}
\usepackage[lf,scale=0.9]{FiraMono}

\title{Fields of The World: A Field Guide for Extracting Agricultural Field Boundaries}

\author{
Isaac Corley\thanks{Corresponding author: \texttt{\scriptsize{isaac@wherobots.com}}}\\
Wherobots\\
\And
Hannah Kerner\\
Arizona State University\\
\And
Caleb Robinson \\
Microsoft AI for Good Research Lab \\
\And
Jennifer Marcus\\
Taylor Geospatial
}

\iclrfinalcopy %
\begin{document}

\maketitle

\begin{abstract}
Field boundary maps are a building block for agricultural data products and support crop monitoring, yield estimation, and disease estimation.
This tutorial presents the Fields of The World (FTW) ecosystem: a benchmark of 1.6M field polygons across 24 countries, pre-trained segmentation models, and command-line inference tools.
We provide two notebooks that cover (1) local-scale field boundary extraction with crop classification and forest loss attribution, and (2) country-scale inference using cloud-optimized data.
We use MOSAIKS random convolutional features and FTW derived field boundaries to map crop type at the field level and report macro F1 scores of 0.65--0.75 for crop type classification with limited labels.
Finally, we show how to explore pre-computed predictions over five countries (4.76M km\textsuperscript{2}), with median predicted field areas from 0.06 ha (Rwanda) to 0.28 ha (Switzerland).
Code and notebooks are available \href{https://github.com/fieldsoftheworld/iclr2026-ml4rs-tutorial}{here}.
\end{abstract}

\section{Introduction}

\begin{wrapfigure}{r}{0.38\textwidth}
  \centering
  \vspace{-10pt}
  \includegraphics[width=0.36\textwidth]{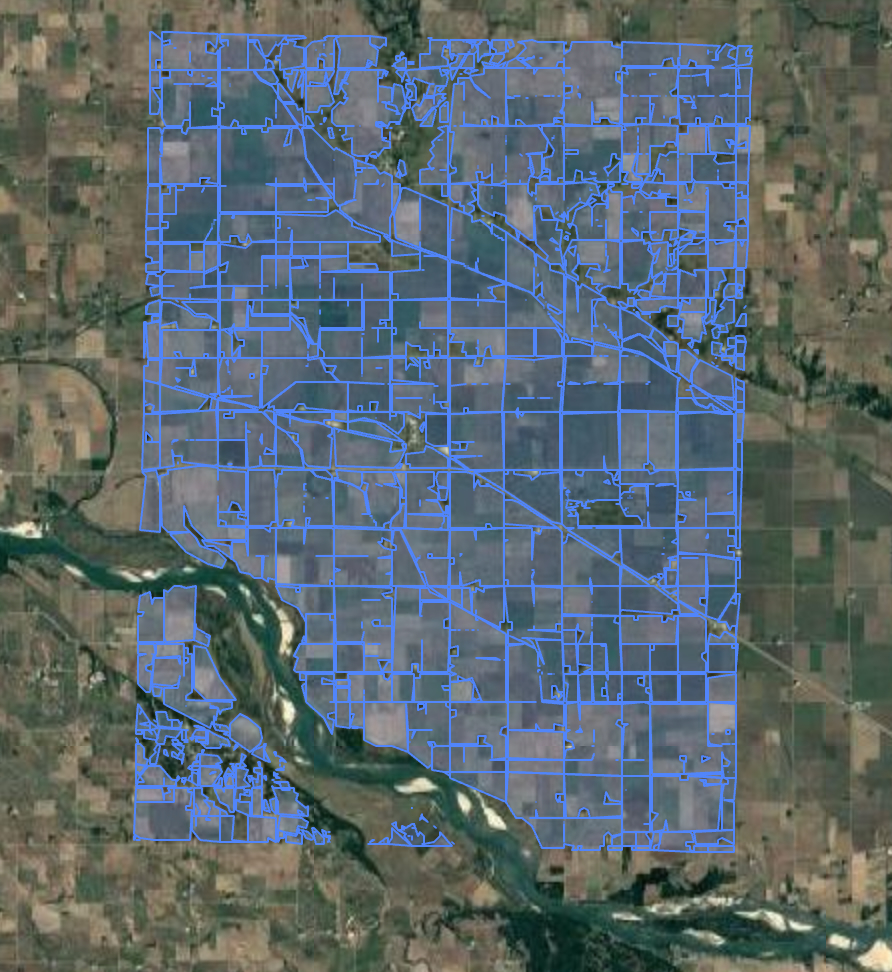}
  \caption{\textbf{FTW field boundary predictions for MGRS tile 14TPN in Iowa, USA.} Blue polygons show extracted agricultural parcels
  overlaid on Sentinel-2 true-color imagery.}\label{fig:polygons}
  \vspace{-20pt}
\end{wrapfigure}

Agricultural monitoring at continental scales informs food security, land-use policy, and compliance frameworks such as the EU Deforestation Regulation (EUDR)~\citep{eudr2023}. Field boundaries define parcels that are the unit of analysis for many agricultural task: crop mapping, yield estimation, subsidy allocation, and deforestation screening (see Figure~\ref{fig:polygons} for and example of field polygon predictions). Further, automated approaches for generating field boundaries from satellite imagery are necessary, as manual digitization of fields is not feasible at scale.

The Fields of The World (FTW) project~\citep{ftw2024} provides a benchmark dataset with 1.6M field polygons across 24 countries, pre-trained segmentation models that use bi-temporal Sentinel-2 imagery as input, and open-source tools for inference and evaluation.
The models are distributed through the \texttt{ftw-tools} package which provides a command line interface for performing inference on Sentinel-2 satellite imagery and postprocessing the result.

This tutorial provides two Python notebooks that implement end-to-end workflows using the \texttt{ftw-tools} package, specifically covering: (1) local-scale field boundary inference used for crop classification and forest loss attribution, and (2) exploration of pre-computed country-scale field boundary outputs and agricultural change detection. Through these workflows, we contribute a reproducible pipeline to download Sentinel-2 imagery and polygonize FTW inference outputs, the integration of these polygons with MOSAIKS embeddings for few-shot crop classification~\citep{rolf2021generalizable}, parcel-level forest loss attribution using Hansen Global Forest Change~\citep{hansen2013high}, and standardized access patterns for pre-computed country-scale predictions stored as cloud-native Zarr archives.

\section{Related Work}

Field boundary extraction has trended from classical image processing and edge detection to deep learning-based modeling~\citep{lavreniuk2025delineate,RUFIN2024104149,waldner2020deep}.
Edge-based and region-growing approaches perform poorly under spectral variability and irregular parcel geometry.
Encoder-decoder architectures (e.g., U-Net) now dominate agricultural segmentation~\cite{estes2024region}, and FTW~\citep{ftw2024} provides a globally diverse training set that supports generalization across different geographies and climate zones.

\begin{figure}[t!]
\centering
\includegraphics[width=0.31\textwidth, valign=c]{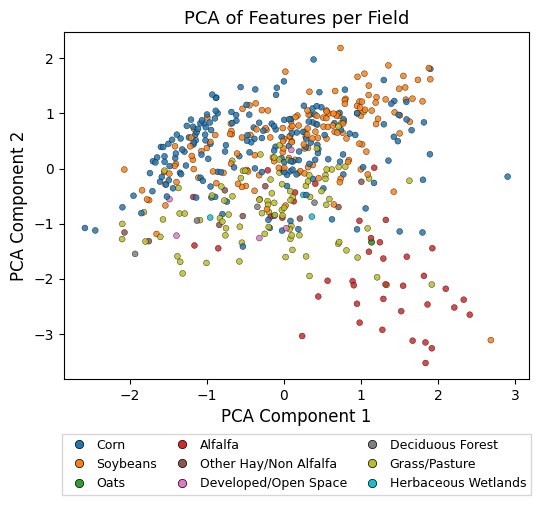}
\includegraphics[width=0.31\textwidth, valign=c]{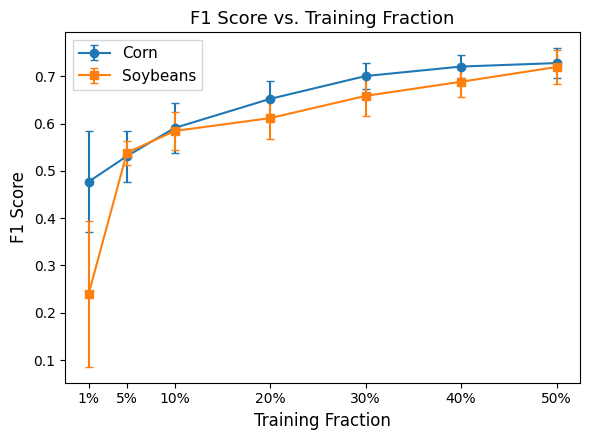}
\includegraphics[width=0.31\textwidth, valign=c]{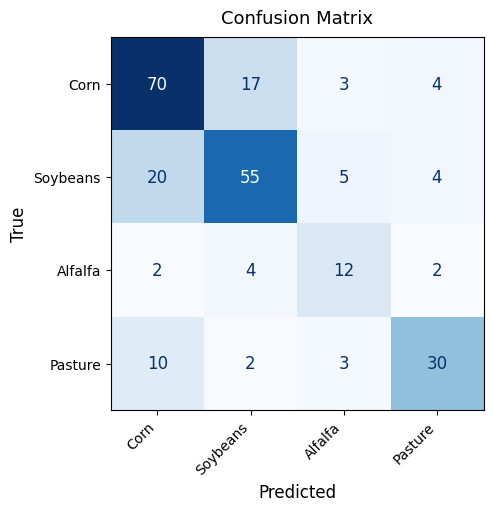}
\caption{\textbf{Crop classification results on Iowa fields}: (\textbf{left}) PCA of 256-dim MOSAIKS embeddings colored by CDL crop type showing separable clusters; (\textbf{center}) macro F1 score versus training data fraction demonstrating few-shot capability; (\textbf{right}) confusion matrix revealing high accuracy for dominant crops (corn, soybeans) with more errors on minority classes.}\label{fig:results}
\end{figure}

\section{Methods}

\paragraph{Data Pipeline.}
Our workflow begins with bi-temporal Sentinel-2 imagery for planting and harvest windows.
We query the Microsoft Planetary Computer STAC catalog~\citep{planetarycomputer} for cloud-free Level-2A surface reflectance scenes.
For each MGRS tile (100$\times$100~km), we define seasonal windows from local crop calendars and compute median composites to reduce atmospheric
artifacts. The MGRS grid provides a standardized tiling scheme for reproducible analysis.

FTW models take two 4-channel inputs (RGB+NIR at 10 m) from early and late season observations. The bi-temporal setup captures
phenological differences between crops and non-agricultural land cover. The acquisition gap is typically 3--6 months, depending
on the growing season.

\paragraph{Field Boundary Extraction.}
The FTW segmentation models use a U-Net model with an ImageNet pretrained EfficientNet-B3 encoder. Training data span 24 countries and a wide range of field sizes ($<$1~ha to $>$100~ha). The model outputs per-pixel probabilities for field interior, field boundary, and background. The \texttt{ftw-tools} package provides a command-line interface for inference:

\begin{lstlisting}
# Install FTW package
pip install ftw-tools

# Download bi-temporal Sentinel-2 imagery and run inference
ftw inference download --win_a <id1> --win_b <id2> --out input.tif --bbox <bbox>
ftw inference run input.tif --out preds.tif --model FTW_PRUE_EFNET_B3_CCBY

# Filter by land cover and convert raster predictions to polygons
ftw inference filter-by-lulc preds.tif --out filtered.tif
ftw inference polygonize filtered.tif --out boundaries.gpkg
\end{lstlisting}

\noindent The post-processing steps threshold softmax outputs at 0.5, apply morphological opening to remove small artifacts, and convert
rasters to polygons via connected components. We optionally filter predictions with land-cover masks (e.g., ESA WorldCover) to
remove forests, water, and urban areas. Figure~\ref{fig:polygons} shows example predictions over Sentinel-2 imagery.

\paragraph{Crop Classification with MOSAIKS.}
For crop mapping we use MOSAIKS~\citep{rolf2021generalizable}, which produces fixed-dimensional embeddings from random convolutional filters, that are shown to have high performance under linear probing in a variety of downstream tasks.

\begin{figure}[t!]
  \centering
  \includegraphics[width=0.98\textwidth]{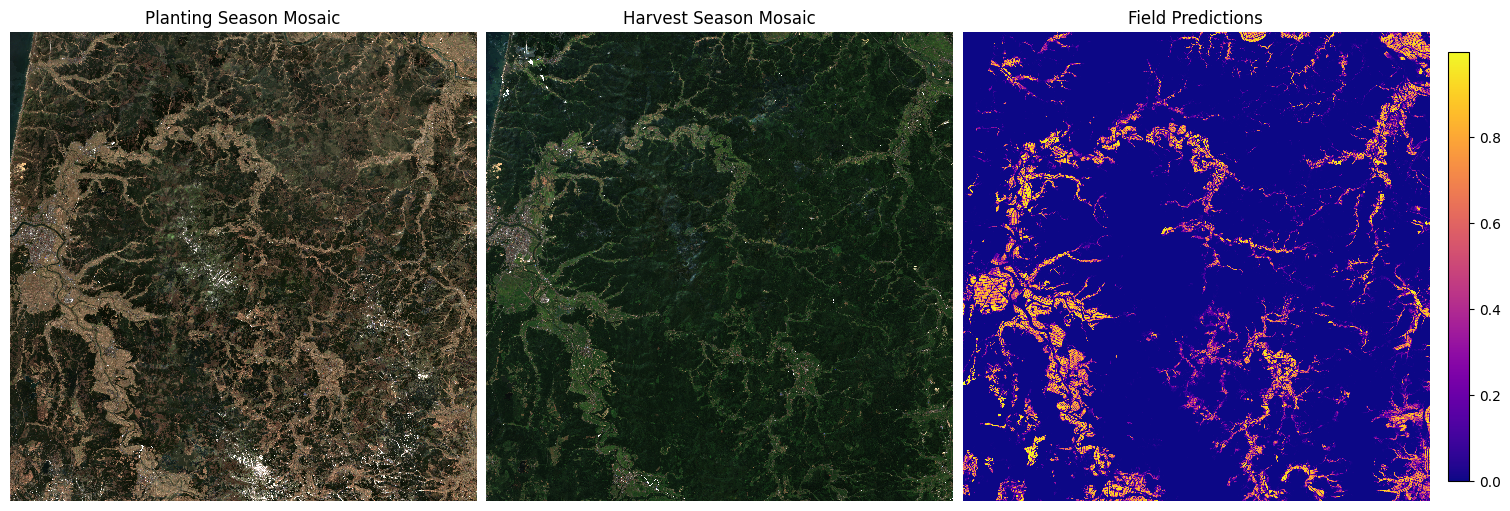}
  \caption{\textbf{Country-scale FTW inference for Japan.} Left: Sentinel-2 RGB mosaic. Right: per-pixel field probability (softmax
  output). Data accessed via cloud-optimized Zarr archives.}\label{fig:japan}
\end{figure}

\begin{wrapfigure}{r}{0.42\textwidth}
  \centering
  \vspace{-10pt}
  \includegraphics[width=0.42\textwidth]{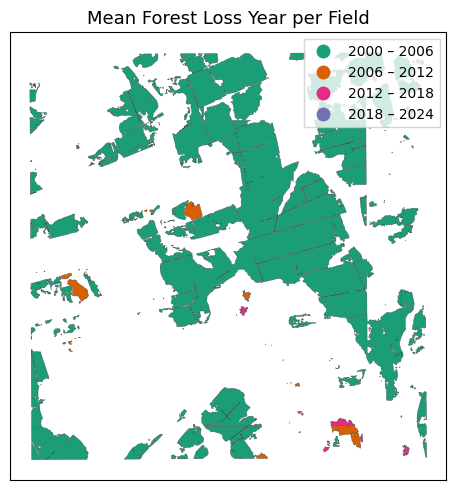}
  \caption{\textbf{Forest loss attribution for MGRS tile 21LXF in Mato Grosso, Brazil.} Colors indicate mean deforestation year per field polygon from Hansen GFC data. Pink and Purple colors denote recent clearing (2012--2024), relevant to EUDR compliance thresholds.}
  \label{fig:forest}
  \vspace{-40pt}
\end{wrapfigure}

For each FTW polygon, we extract the corresponding Sentinel-2 pixels, apply 256 random 3$\times$3 filters (Gaussian initialization), apply ReLU, and
compute a masked global average pooling to obtain a 256-dim embedding.
We train a logistic regression classifier using Cropland Data Layer (CDL) from the USDA~\citep{cdl} as labels.

\paragraph{Forest Loss Monitoring.}
We integrate Hansen Global Forest Change (GFC)~\citep{hansen2013high} to attribute forest loss to field polygons.
GFC provides annual tree cover loss predictions at 30 m/px spatial resolution from 2000 to 2023.
We spatially join FTW derived field boundary polygons with loss-year rasters from GFC and compute zonal statistics (mean, min, max loss year) per field, which identifies parcels with recent deforestation, which is relevant to measuring EUDR compliance after 2020.

\paragraph{Country-Scale Inference.}
We provide pre-computed predictions for five countries (Japan, Mexico, Rwanda, South Africa, Switzerland) totaling 4.76M~km\textsuperscript{2} across two seasons and two years (2023--2024). Seasonally-aligned mosaics are constructed using latitude-dependent heuristics to identify optimal planting and harvest windows; a greedy algorithm selects cloud-free Sentinel-2 scenes to maximize coverage. Inference uses a sliding window of 256$\times$256 pixels with 25\% overlap. Gaussian-weighted averaging (apodization) merges overlapping patches, suppressing lower-quality edge predictions and reducing tiling artifacts.
Outputs are reprojected to Web Mercator (EPSG:3857) on-the-fly and written in parallel to cloud-optimized Zarr archives, enabling lazy loading and spatial queries without full downloads. Median predicted field areas range from 0.06~ha (Rwanda) to 0.28~ha (Switzerland), reflecting variation between smallholder and mechanized systems.

\section{Results}

\paragraph{Crop Classification.}
We evaluate crop classification on MGRS tile 14TPN (Iowa, USA) using FTW polygons and the following CDL labels: corn, soybeans, grass/pasture, and alfalfa. Logistic regression on MOSAIKS embeddings yields macro F1 scores of 0.65--0.75 depending on training size (Figure~\ref{fig:results}). PCA of embeddings shows separable clusters by crop type.

Using 10\% of labeled fields for training yields F1 scores within 5\% of full-data training. Figure~\ref{fig:results}c shows high accuracy for dominant classes (corn, soybeans; F1 $>$ 0.8) and more confusion among minority classes, consistent with class imbalance.

\paragraph{Forest Loss Attribution.}
Applying the workflow to Brazil (MGRS tile 21LXF, Mato Grosso) links GFC loss years to individual fields (Figure~\ref{fig:forest}). Recent deforestation (2015--2020) appears in warmer colors.

\paragraph{Country-Scale Predictions.}
The Zarr archives store both Sentinel-2 mosaics (RGB+NIR at 10 m) and raw softmax probabilities for background, field interior, and field boundary classes. Users can query arbitrary spatial extents without downloading complete datasets. Figure~\ref{fig:japan} shows country-scale predictions for Japan with input imagery and output probability maps. The raw probabilities support custom thresholding, watershedding, or vectorization beyond the defaults provided in the \texttt{ftw-tools} command line interface.

\begin{figure}[t!]
  \centering
  \includegraphics[width=0.98\textwidth]{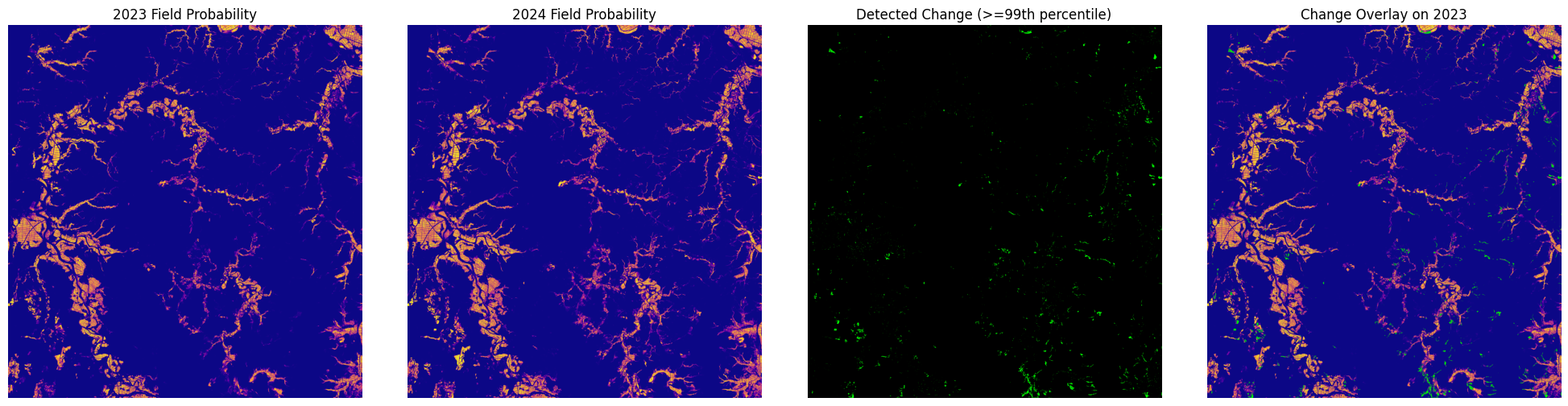}
  \caption{\textbf{Year-over-year change detection from FTW field probability maps.} Panels show 2023 and 2024 field probability, detected
  changes (\(\ge\)99th percentile), and a change overlay on the 2023 map.}\label{fig:change}
\end{figure}

We also compute year-over-year change by differencing field probability maps from 2023 and 2024 (Figure~\ref{fig:change}). Pixels exceeding the 99th percentile of absolute difference are flagged as potential agricultural expansion or abandonment, demonstrating how the Zarr archives enable temporal analysis without rerunning inference.

\section{Conclusion}

This tutorial demonstrates how the Fields of The World ecosystem supports end-to-end agricultural monitoring workflows. We show that FTW segmentation models generalize across diverse regions and that the extracted polygons integrate with downstream tasks including crop classification via MOSAIKS embeddings and forest loss attribution via Global Forest Watch labels. Pre-computed country-scale predictions stored as cloud-optimized Zarr enable efficient spatial queries and temporal change detection without local reprocessing and can be openly used for a variety of downstream applications.

\bibliography{iclr2026_conference}
\bibliographystyle{iclr2026_conference}

\end{document}